\documentclass{article}

\usepackage{arxiv}

\usepackage{floatrow}   
\usepackage{subcaption}
\usepackage{graphicx} 
\usepackage{amsmath}
\usepackage{natbib}
\usepackage{algpseudocode,algorithm,algorithmicx}

\author{
    Beren Millidge \\
    Department of Informatics \\
    University of Edinburgh \\
    United Kingdom
}
\title{Deep Active Inference as Variational Policy Gradients}

\begin{document}

\maketitle

\begin{abstract}

Active Inference is a theory of action arising from neuroscience which casts action and planning as a bayesian inference problem to be solved by minimizing a single quantity -- the variational free energy. Active Inference promises a unifying account of action and perception coupled with a biologically plausible process theory. Despite these potential advantages, current implementations of Active Inference can only handle small, discrete policy and state-spaces and typically require the environmental dynamics to be known. In this paper we propose a novel deep Active Inference algorithm which approximates key densities using deep neural networks as flexible function approximators, which enables Active Inference to scale to significantly larger and more complex tasks. We demonstrate our approach on a suite of OpenAIGym benchmark tasks and obtain performance comparable with common reinforcement learning baselines. Moreover, our algorithm shows similarities with maximum entropy reinforcement learning and the policy gradients algorithm, which reveals interesting connections between the Active Inference framework and reinforcement learning.

\keywords{Active Inference, Predictive Processing, OpenAI Gym, Neural Networks, Policy Gradients, Actor-Critic, Reinforcement Learning}
\end{abstract}
\bigskip

Active Inference is a proposed unifying theory of action and perception emerging out of the Predictive Coding \citep*{clark2013whatever,clark2012dreaming,clark2015surfing,rao1999predictive,friston2003learning} and Bayesian Brain \citep*{knill2004bayesian, doya2007bayesian,friston2012history} theories of brain function. \citep*{friston2009reinforcement,brown2011active,adams2013predictions}. It has been applied in a variety of ways including in modelling choice tasks \citep*{friston2013anatomy,friston2014anatomy}, serving as a basis for exploration, artificial curiosity \citep*{friston2015active,friston2017active}, the explore-exploit trade-off \citep*{friston2012active}, and potentially for illuminating neuropsychiatric disorders \citep*{mirza2019impulsivity,adams2012smooth,barrett2016active}. Moreover, a neuroscientifically grounded process theory has been developed \citep*{friston2017active}, based on variational message passing \citep*{parr2018generalised,parr2019neuronal,van2019simulating,friston2017graphical}, that can replicate several observed neuropsychological processes such as repetition suppression, mismatch negativity, and perhaps even place-cell activity. \citep*{friston2017active}.

Active Inference casts action and perception as bayesian inference problems which can both be solved simultaneously through a variational approach that minimizes a single quantity -- the variational free-energy. This is in line with the Free-Energy Principle: a deeper theory emerging from predictive processing \citep*{friston2010free,friston2006free,friston2007free,friston2009free,friston2019free} which proposes that the brain, and perhaps all far-from-equilibrium self-organising systems must in some sense minimize their free-energy \citep*{friston2012free,karl2012free}. Despite the free-energy being a rather esoteric concept, under certain assumptions the free-energy principle can be translated into a biologically and neuroscientifically plausible process theory \citep*{friston2003learning, friston2005theory} that could theoretically be implemented in the brain. \footnote{Some tutorial introductions to the Free-Energy Principle and its applications to neuroscience are: \citet*{bogacz2017tutorial,buckley2017free,millidge2019combining}.} The core idea is that the brain possesses hierarchical generative models capable of generating expected sense-data, which are trained by minimizing the prediction error between the predicted and observed sense-data. Prediction error thus becomes a general unsupervised training signal, used to successively infer and improve our understanding of the state of the world. Due to the emphasis on prediction error, this general theory is known as Predictive Processing. Active Inference extends this idea by applying it to action. There are two ways to minimize prediction error. The first is to update internal models to accurately account for incoming sense-data. This is perception. The second is to take actions in the world so as to bring the incoming sense-data into agreement with the prior predictions. This is action. This duality lets Active Inference treat action and perception under the same formalism, and enables both to be optimized simultaneously by minimizing the variational free-energy.

Active Inference was first applied to continuous time, state, and action spaces. \citep*{friston2009reinforcement}. A discrete version was later developed which is more in line with current trends in reinforcement learning and optimal control \citep*{friston2012active}, which we focus on here. Although there are significant differences from paper to paper, and we have elided much detail, the general setup of discrete Active Inference is as follows:

There is an agent which exists in a Partially Observed Markov Decision Process (POMDP). The agent receives observations \textit{$o$} from an environment which has hidden states \textit{$s$} which the agent tries to infer. The agent can also take actions which change the environment's state. The agent's "goal" is to minimize its expected free energy into the future up to some time horizon T. The agent then infers its own actions at the current time to be consistent with this goal.

The agent is equipped with a generative model of its observations, states, and actions which can be factorized as follows:
\begin{align}
p(o,s,a,\gamma) = p(o|s)p(s|s_{t-1},a_{t-1})p(a|\gamma)p(\gamma)
\end{align}
where $\gamma$ is a precision parameter which affects the distribution of actions. Each of the distributions in the factorized generative model is typically represented as a single matrix which is usually provided by the experimenter rather than being learned from experience. \footnote{Several papers \citep*{friston2016active,schwartenbeck2019computational} do learn at least the "A" matrix representing $p(o|s)$ which can be done by setting hyperparameters governing the distribution of the values in the A matrix and then deriving additional variational update rules for these hyperparameters, but this is not the norm and it has only been applied to learn the "A" matrix.}

The agent then inverts this generative model through a process of approximate variational inference. This works by defining variational distributions $Q(s,a)=Q(s|\hat{s})Q(a|\hat{\pi})$  minimizing the  KL-divergence between these distributions and the generative model. The divergence between the variational distribution and the generative model is called the variational free energy. 

To infer its policy, the agent needs to compute the expected-free-energy (EFE) of each policy, which is simply the sum of the free energies expected under the variational posterior up to the time horizon. What this means in practice is that, for every possible policy, the agent needs to run forward its generative model from the current time until the time horizon, generating fictitious future states and observations it projects it will be in if it follows that policy. It them must compute the free-energy of those states and observations and add them all up to get an estimate of the value of any particular policy. The agent can then sample actions from its action posterior using a Boltzmann distribution with the $\gamma$ parameter acting as an inverse temperature.

Due to the need to enumerate over every possible policy and project them forwards in time up until the time horizon, this algorithm quickly becomes intractable for large policy spaces or time horizons. It also has trouble representing large state-spaces. We refer to this type of algorithm as tabular Active Inference, by analogy to tabular reinforcement learning, which represents every state in the state-space explicitly as a giant table, and runs into similar scaling issues \citep*{kaelbling1996reinforcement,sutton1998introduction}. Because of these scaling issues, tabular Active Inference has not been applied to any non-toy tasks with large state or action spaces. 

In this paper, inspired by recent advances in machine learning and variational inference \citep*{goodfellow2016deep,kingma2013auto}, we propose a novel deep Active Inference algorithm which uses deep neural networks to approximate the key densities of the factorized generative model. This approach enables Active Inference to be scaled up to tasks significantly larger and more complex than any attempted before in the tabular Active Inference literature, and we demonstrate performance comparable to common reinforcement learning algorithms for several baseline tasks in OpenAIGym \citep*{brockman2016openai} -- the CartPole, Acrobot, and Lunar-Lander task. Our algorithm does not need pre-specified transition or observation models hard-coded into the algorithm as it can learn flexible nonlinear functions to approximate these densities which can be optimized purely through a gradient descent on the variational free-energy functional without needing hand-crafted variational update rules. Moreover, we show how one can use a bootstrapping estimation technique to obtain amortized estimates of the expected-free-energy for a state-action pair without needing to explicitly project the policy forward through time, which potentially enables the algorithm to handle long or infinite time horizons.

We find that the mathematical form of the policy-selection part of our algorithm is somewhat similar to the policy gradients and actor-critic algorithms in reinforcement learning, despite having been derived from completely different frameworks and objectives. We compare and contrast these algorithms with our own algorithm, and highlight how Active Inference natively includes several adjustments that have been empirically found to improve policy gradients but which fall naturally out of our framework. We also investigate how Active Inference includes exploration and information-gain terms and we compare them to related work in the reinforcement learning literature.

\section{Deep Active Inference}

Our deep Active Inference algorithm uses the same POMDP formulation as the tabular version. There is an environment which has internal states which then generate observations which the agent receives. The agent can then take actions in response to these observations that affect the internal state of the environment, and thus the observations that the agent receives in the future. In our case, the agent also receives rewards from the environment, which it uses to construct better policies. The states of the environment are Markov, which means that the next state depends only on the current state and the agent's action. The observations the environment generates are not necessarily Markov. 

The agent maintains a generative model of the environment with the same high-level structure comprising states, observations, actions, and rewards. Unlike the observations, however, the states and actions are hidden or latent variables, which the agent must infer. In bayesian terms, this means that the agent must compute the following posterior probability, where $s_{<t}$ and $a_{<t}$ are the histories of previous states and actions up to the current time t.

\begin{align*}
p(s,a | o, s_{<t}, a_{<t})
\end{align*}

Directly computing this posterior through bayesian inference is intractable for any but the simplest cases. Instead variational methods are used. These posit additional variational densities Q that the agent controls, which are then optimized by minimizing the KL divergence between them and the true posterior so that ultimately the Q densities approximate the true posterior densities. The variational density to minimize is thus:
\begin{align}
KL[Q(s,a)||p(s,a, | o, s_{<t}, a_{<t})] = \int Q(s,a) log(\frac{Q(s,a)}{p(s,a, | o, s_{<t}, a_{<t}))})
\end{align}
Using bayes' rule, the properties of logarithms, and the linearity of the integral, we can split this expression up as follows:
\begin{align}
KL[Q(s,a)||p(s,a, | o, s_{<t}, a_{<t})]
&= E_{Q(s,a)}[ logQ(s,a)] + E_{Q(s,a)}[logp(s,a,o,s_{<t},a_{<t})] + \int Q(s,a)logp(o, s_{<t}, a_{<t})\\
 &= \int Q(s,a) log(Q(s,a) + \int Q(s,a)logp(s,a,o,s_{<t},a_{<t}) + logp(o,s_{<t}, a_{<t})
\end{align}
This last integral vanishes since $logp(s,a,o,s_{<t},a_{<t})$ has no dependence on any of the variables in Q, so these can be taken out of the integral, and as the remainder is just a distribution it integrates to  1. We can thus see that:
\begin{align}
KL[Q(s,a)||p(s,a, | o, s_{<t}, a_{<t})] = KL[Q(s,a)||p(s,a,o,s_{<t},a_{<t})] + logp(o,s_{<t},a_{<t})
\end{align}
Since the \textit{logp} term does not depend on the parameters of Q (and since KL divergence is non-negative), minimizing the KL divergence on the right-hand side is the same as minimizing that on the left. This then replaces the difficulties of minimizing the KL between the variational distribution and the true posterior (which is unknown), with that of the KL divergence between the variational distribution and the joint distribution, which is given by the generative model. The variational free energy is simply the KL divergence between the variational and joint distributions:
\begin{align}
KL[Q(s,a)||p(s,a,o,s_{<t},a_{<t})] = F
\end{align}

This quantity is also the same as the evidence-based-lower-bound ELBO used in machine learning and variational inference \citep*{hoffman2013stochastic,blei2017variational}. We now have to deal with the free energy term, and specifically the joint distribution $p(s,a,o,s_{<t},a_{<t})$. This term can be factorized according to the generative model of the agent. The agent's generative model assumes that the agent exists inside a POMDP such that the generative process is that observations depend on states, states depend on the previous state and the previous action, and the current action is inferred from the current state only. These assumptions result in the following factorization for the joint density:
\begin{align}
p(s,a,o,s_{<t},a_{<t}) = p(o|s)p(a|s)p(s|s_{t-1}, a_{t-1}) = F
\end{align}

The "history terms" $s_{<t}$ and $a_{<t}$ are replaced by $s_{t-1}$ and $a_{t-1}$ because of the Markov assumption on the states and actions. The variational distribution is taken to factorize as $Q(s,a) = Q(a|s)Q(s)$. This factorization does not impose any additional assumptions and follows directly from the laws of probability.. The free energy to be minimized is thus:
\begin{align}
F = KL[Q(a|s)Q(s))||p(o|s)p(a|s)p(s|s_{t-1}, a_{t-1})] 
\end{align}
Using standard properties of logs and the definition of the KL divergence, the expression for the free energy splits apart into three terms:
\begin{align}
-F =\int Q(s)logp(o|s) + KL(Q(s)||p(s|s_{t-1},a_{t-1})] + E_{Q(s)}[KL[Q(a|s)||p(a|s)]
\end{align}

The key element of our method is using deep neural networks to approximate each of the key densities in this expression. Looking at the first term, we see we need to approximate the densities \textit{Q(s)} and $logp(s|o)$. These two densities, one mapping from observations to states and the other mapping back from states to observations is highly reminiscent of the variational-autoencoder (VAE) objective, \citep*{kingma2013auto} and can be modelled directly as one. \footnote{For a tutorial on VAEs see \citet*{doersch2016tutorial}. }

The second term is the divergence between the posterior expected given the observation and the prior expected value of the state given the previous state and action. The variational posterior is given by the encoder model of the observation model, while the prior $p(s|s_{t-1},a_{t-1})$ can be modelled, under Gaussian assumptions, directly by a neural network which outputs the mean and variance of a Gaussian given the previous state and action. In this paper we use a simple feedforward network for this, but more complex stateful models such as LSTMs could also be used. Assuming both the posterior and prior densities are Gaussian, the KL divergence is computable analytically, so there is no computational difficulty here. These two terms and their instantiations as neural networks take care of the perception aspect of the Active Inference agent.

The key term for Active Inference is the third term. Let's examine it in more detail. We can decompose the KL divergence into an energy and an entropy term:
\begin{align}
KL[Q(a|s)||p(a|s)] = \int Q(a|s) logp(a|s) + H(Q(a|s))
\end{align}
The variational density $Q(a|s)$ is under complete control of the agent, and we parametrize it by a deep neural network. This makes the entropy term $H(Q(a|s))$ simple to compute as it is simply the entropy of the action distribution output by the neural network which can be computed simply as a sum for discrete actions. The energy term is more tricky. This is because it involves the true action posterior $p(a|s)$, which we do not know precisely. First, we will make some assumptions about the form of this density. Specifically, we will assume, following \citet*{friston2015active} and \citet*{schwartenbeck2019computational}, and as in the tabular case, that the agent expects that it will act to minimize its expected-free-energy into the future, and that the distribution over actions is a precision-weighted Boltzmann distribution over the expected free energies. That is:
\begin{align}
p(a|s) = \sigma(- \gamma G(s,a))
\end{align}
Where $\sigma$ is a softmax function. This just tells us that the "optimal" free-energy agent would first compute the expected free-energy of all paths into the future for each action it could take, and then choose an action probabilistically by sampling from a Boltzmann distribution over the expected free energies for each action. This method has some empirical support. Boltzmann or softmax choice rules have been regularly used to model decision-making in humans and other animals \citep*{gershman2018deconstructing,gershman2018uncertainty,daw2006cortical} and similar rules are applied in reinforcement learning under the term Boltzmann Exploration \citep*{cesa2017Boltzmann}. The key term in this equation is the expected free energy $G(s,a)$. This is a path integral (or sum) of the free-energy of the expected trajectories into the future given the current state and action. This means that in the case of discrete time-steps and an ultimate time horizon T, we can write this as a simple sum:
\begin{align}
G(s,a) = \sum_t^T G(s_t, a_t)
\end{align}
We first take out the first term obtain an expression of the expected free energy for a single time-step:
\begin{align}
G(s,a) = G(s_t,a_t) + E_{Q(s_{t+1},a_{t+1})}[\sum_{t+1}^TG(s_{t+1},a_{t+1})]
\end{align}
We can then expand the expected free energy for a single time-step using the definition of the free energy from before as the KL divergence between the variational distribution and the joint distribution to obtain \footnote{Action is not included as a free parameter in the variational or generative densities since the expected-free-energy is a function of both states and actions -- i.e. the EFE is evaluated for every action so the action is implicitly conditioned on).}:
\begin{align}
G(s_t,a_t) &= KL[Q(s)||p(s,o)]\\
&= \int Q(s)[logQ(s) - logp(s,o)] \\
&= \int Q(s)[logQ(s) - logp(s|o) - logp(o)] \\
&= \int Q(s)[logQ(s) - logQ(s|o) - logp(o)] \\
&= -logp(o) + \int Q(s)[logQ(s) - logQ(s|o)] \\
&= -r(o) +  \int Q(s)[logQ(s) - logQ(s|o)]
\end{align}
In the third line, since we do not know the true posterior distribution of the states given the observations in the future, we approximate it with our variational density. In the penultimate line, $logp(o)$ term can be taken out of expectation since it has no dependence on \textit{s}. This term is crucial to Active Inference as it is the prior expectation over outcomes, which encodes the agents preferences about the world. In the final line we replace the prior probability of an observation \textit{-logp(o)} directly with the reward \textit{-r(o)}. This is because, in Active Inference, the agent is simply driven to minimize surprise, and therefore all goals must be encoded as priors built into the agent's generative model, so that the least surprising thing for it to do would be to achieve its goals. Due to the complete class theorem \citep*{friston2012active} any scalar reward signal can be encoded directly as a prior using $p(o) \propto exp(r(o))$. In this paper,to enable effective comparisons with reinforcement learning methods, the agent's priors simply are to maximize rewards. However, it is important to note that the Active Inference framework is actually more general than reinforcement learning. It can represent reward functions directly as priors using the complete class theorem, but it can also encode other more flexible functions. Additionally, the action posterior $p(a|s)$ does not necessarily have to be computed using the expected-free-energy. For instance, the action probabilities could be provided directly by observing another agent, which would allow the Active Inference agent to switch seamlessly between imitation and reinforcement learning styles.

The prior term represents the external reward the agent receives, but there is a second term $\int Q(s)[logQ(s) - logQ(s|o)]$ which represents something quite different. It requires the maximization of the difference between the prior over future states, and the future "posterior" generated after "observing" the fictive predicted future outcome. This incentivizes visiting states where either the transition or observation model are poor, thus endowing the agent with what can be thought of as an intrinsic curiosity. In the Active Inference literature, this term is often called the epistemic or intrinsic value \citep*{friston2015active}.

We can then represent the total expected free energy as the sum:
\begin{align}
G(s,a) = -r(o) +  \int Q(s)[logQ(s) - logQ(s|o)] + E_{Q(s_{t+1},a_{t+1})}[\sum_{t+1}^TG(s_{t+1},a_{t+1})]
\end{align}

Trying to compute this quantity exactly is intractable due to the need to explicitly compute many future trajectories and the expected-free-energy associated with each one. However, it is possible to learn a bootstrapped estimate of this function from samples using a neural network to learn an amortized inference distribution. We define an approximate expected-free-energy (EFE)-value network $G_\phi(s,a)$, with parameters $\phi$, which  maps a state action pair to an estimated EFE. This estimated EFE is then compared to a second estimate of the EFE $\hat{G(s,a)}$ which uses the free energy calculated at the current time-step but approximates the rest of the trajectory with another estimate by the EFE-value net estimate for the next time-step. That is:
\begin{align}
\hat{G(s,a)} =  -r(o) +  \int Q(s)[logQ(s) - logQ(s|o)] + G_\phi(s,a)
\end{align}

The difference between the two estimates can then be minimized by a gradient descent procedure with respect to the parameters of the EFE-valuenet $\phi$. In this paper the L2 norm is used as a loss function for the gradient descent:
\begin{align}
L = ||G_\phi(s,a) - \hat{G(s,a)}||^2
\end{align}

This procedure is analogous to bootstrapped value or Q function estimation procedures in reinforcement learning, for which guarantees of convergence exist in tabular cases. It also empirically has been found to work for deep neural networks in practice, albeit with various techniques needed to boost the stability of the optimization procedure, despite the lack of any theoretical convergence guarantees.

Given that we now possess a means to estimate \textit{G(s,a)} and thus the action posterior $p(a|s)$, the action model $Q(a|s)$ can be trained to directly minimize the loss function $L = \int Q(a|s)logp(a|s) + H(Q(a|s))$. Gradients of this expression with respect to the parameters of $Q(a|s)$ can be computed analytically or by using automatic differentiation software.

To recap, the deep Active Inference agent possesses four internal neural networks. A perception model which maps observations to states and back again and models the distributions $Q(s|o)$ and $p(o|s)$, and is trained with a VAE-like log-probability-of-observations loss. A transition model which models the distributions $p(s_t|s_{t-1},a_{t-1})$ is trained to minimize differences between the predicted transition and the actually occurring state at the next time-step obtained through $Q(s_t|o_t)$. An action model, which models the distribution $Q(a_t|s_t)$, and can be trained directly through gradient descent on the loss function  $L = \int Q(a|s)logp(a|s) + H(Q(a|s))$, and a value network $G_\phi(s,a)$ that is trained through a bootstrapped estimate of the expected-free-energy, as explained above.

We present our deep-active-inference algorithm in full below:

\begin{algorithm}[H]
  \caption{ \textbf{Deep Active Inference}}
  \begin{algorithmic}
    \Statex
    \Statex \textbf{Initialization:}
    \Statex Initialize Observation Networks $Q_\theta(s|o), p_\theta(o|s)$ with parameters $\theta$.
    \Statex Initialize State Transition Network $p_\phi(s|s_{t-1},a_{t-1})$ with parameters $\phi$ 
    \Statex Initialize policy network $Q_\xi(a|s)$ with parameters $\xi$
    \Statex Initialize bootstrapped EFE-network $G_\psi(s,a)$ with parameters $\psi$ 
    \Statex Receive prior state $s_0$
    \Statex Take prior action $a_0$
    \Statex Receive initial observation $o_1$
    \Statex Receive initial reward $r_1$
    \Function{Action-Perception-Loop}{}
        \While{$t < T$ }
            \State $\hat{s_t} \leftarrow Q_\theta(s|o)(o_t)$
            \Comment{Infer the expected state from the observation}
            \State $\widehat{s_{t+1}} \leftarrow p_\phi(s|s_{t-1},a_{t-1})(\hat{s})$
            \Comment{Predict the state distribution for the next time-step}
            \State $a_t \sim  Q_\xi(a|s)$
            \Comment{Sample an action from the policy and take it}
            \State Receive observation $o_{t+1}$
            \State Receive reward $r_{t+1}$
            \State $\hat{s_{t+1}} \leftarrow Q_\theta(s|o)(o_{t+1})$
            \Comment{Infer expected state from next observation}
            \State Compute the bootstrapped EFE estimate of from the current state and action: 
            \State $\widehat{G(s,a)} \leftarrow r_{t+1} + E_{Q(s_{t+1}}[log\widehat{s_{t+1}} - log\hat{s_{t+1}}] + E_{Q(s_{t+1},a_{t+1})}[G_\psi(s_{t+2},a_{t+2})]$
            \State Compute the Variational Free Energy F:
            \State $F \leftarrow E_{Q(s)}[logp(o|s)] + KL[\hat{s_{t+1}}||\widehat{s_{t+1}}] + E_{Q(s)}[\int da Q_\xi(a|s)\sigma(-\gamma G_\psi(s,a)(s_{t+1})) + H(Q_\xi(a|s))]$
        
            \State $\theta \leftarrow \theta + \alpha \frac{dF}{d\theta}$
            \Comment{Update the $\theta$ parameters}
            \State $\phi \leftarrow \phi + \alpha \frac{dF}{d\phi}$
            \Comment{Update the $\phi$ parameters}
            \State $\xi \leftarrow \xi + \alpha \frac{dF}{d\xi}$
            \Comment{Update the $\xi$ parameters}
            \State $L \leftarrow ||\widehat{G(s,a)} - G_\psi(s,a)||^2$
            \Comment{Compute the boostrapping loss}
            \State $\psi \leftarrow \psi + \alpha \frac{dL}{d\psi}$
            \Comment{Update the $\psi$ parameters}
     \EndWhile
    \EndFunction
    \end{algorithmic}
\end{algorithm}

Unlike the tabular Active Inference algorithms proposed by Friston and colleagues, this algorithm approximates all important densities with neural networks, which can all be optimized through a simple gradient descent procedure on the expression for the variational free energy. The computation graph is fully differentiable so that derivatives of the parameters of the networks can be computed automatically using automatic differentiation software without the need for hand-derived complex variational update rules or black-box optimization techniques. Moreover, although in this paper the densities were approximated using simple multi-layer perception networks, in principle each density can be approximated by a neural network, or other differentiable function, of any size or complexity, thus enabling this algorithm to scale indefinitely.

\section{Relation to Policy Gradients}

Reinforcement learning is perhaps the dominant paradigm used to train agents to solve complex tasks with high dimensional state-spaces \citep*{mnih2015human,lillicrap2015continuous,silver2017mastering}. Reinforcement learning also formulates the action-problem as an MDP with states, actions and rewards. In reinforcement learning, however, instead of acting to minimize expected surprise, we simply maximize expected rewards. The agent's goal at every time-step is simply to maximize the sum of discounted rewards over its trajectories into the future. This can be written as:
\begin{align}
G_t = \sum_i^T \gamma^{i-1}r(s_t,a_t)
\end{align}
Where $\gamma$ is a discount factor that reduces the impact of future rewards. We can also define state and state-action reward functions which map states and actions to the reward expected from that state or state-action pair under a particular policy.
\begin{align}
Q^\pi(s,a) = E_\pi[G_t | s=s,a=a] \\
V^\pi(s) = E_\pi [E_a[G_t | s=s]]
\end{align}

The goal of an agent is to find a policy which can maximize the expected sum of discounted rewards. This goal can be written as the objective function:
\begin{align}
J(\theta) = E[G_t] = \sum_{t=0}^{t=\infty} \int p(s_t | s_{t-1},a_{t-1})p_\theta(a_{t-1}|s_{t-1}) G(s_t,a_t)ds da
\end{align}

Where $\theta$ are the parameters of the policy which outputs the distribution $p_\theta(a|s)$. There are two ways to optimize this objective. The first is to obtain it directly by maximizing over the Q function, and thus not explicitly represent the policy at all. This leads to the Q and TD learning family of algorithms. The second way is to explicitly represent the policy, for instance using a deep neural network, and fit it directly. We focus on this second approach since it bears the greatest similarities with our deep Active Inference algorithm.

We can compute the gradients of \textit{J} with respect to $\theta$ directly using the following log-gradient trick \citep*{sutton2000policy}:
\begin{align}
\nabla_\theta J(\theta) &= \sum_t \int \nabla_\theta p(s_t | s_{t-1},a_{t-1})p_\theta(a_{t-1}|s_{t-1})G(s_t,a_t) \\
&= p(s_t | s_{t-1},a_{t-1})p_\theta(a_{t-1}|s_{t-1}) \frac{\nabla_\theta p(s_t | s_{t-1},a_{t-1})p_\theta(a_{t-1}|s_{t-1})}{p(s_t | s_{t-1},a_{t-1})p_\theta(a_{t-1}|s_{t-1})}G(s_t,a_t) \\
&= \sum_t \int p(s_t | s_{t-1},a_{t-1}) \nabla_\theta logp_\theta(a_{t-1}|s_{t-1}) G(s_t,a_t) \\
&= \sum_t E_{p(s_t | s_{t-1},a_{t-1})}[\nabla_\theta logp_\theta(a_{t-1}|s_{t-1}) G(s_t,a_t)]
\end{align}

These algorithms directly maximize the reinforcement objective and so, unlike Q-learning methods, have some guarantees of formal convergence. However, the gradients often suffer from high variance. The return $G(s_t,a_t)$ can be estimated using Monte-Carlo methods or can be approximated using a function approximation method such as Q-learning. Algorithms that do the latter are called actor-critic algorithms since they contain both an "actor" network - the policy, and a "critic" network which learns to approximate the value function.

Of interest is the close similarity between our algorithm which was derived directly from the variational free-energy using an inference procedure, and the policy gradient updates derived from maximizing the total discounted sum of expected returns. We write the out the loss functions side by side, while simplifying some of the extraneous notation to make the comparison more clear. $J_{AI}$ is Active Inference and $J_{PG}$ represents policy gradients.
\begin{align}
J_{AI}(\theta) &= E_{Q(s)}[\int da Q_\theta(a|s)logp(a|s) + H(Q(a|s))] \\
J_{PG}(\theta) &= E_{p(s,a)}[logp_\theta(a|s)G(s,a)]
\end{align}

There are several interesting similarities and differences. The first is the additional entropy term $H(Q(a|s))$ in the Active Inference objective. This means that the Active Inference agent does not merely try to maximize the expected-free-energy into the future, it does so while also trying to maximize the entropy of the distribution over its actions. Essentially, this maximum entropy objective makes the agent try to act as randomly as possible while still achieving high reward, which significantly aids exploration. Interestingly, a recent strand of reinforcement learning literature also focuses on adding an entropy regularization term to the loss \citep*{haarnoja2017reinforcement,haarnoja2018acquiring,ziebart2008maximum,rawlik2010approximate,rawlik2013stochastic,haarnoja2018soft} and have shown this to empirically aid performance and exploration on many benchmark tasks. The maximum-entropy framework has also inspired work on relating reinforcement learning to probabilistic and variational approaches\citep*{levine2018reinforcement,fellows2018virel}. 

The second difference between the algorithms lies in the value function. Policy gradients uses the state-action value functions directly, while Active Inference replaces this with a log probability which is derived from a precision-weighted softmax over the value-function. It is unclear which approach is to be preferred, although the log-probabilities may help reduce the variance of the gradient by reducing the magnitude of the multiplier since probabilities are inherently normalized. This difference is central to Active Inference, which at a high level aims to minimize surprise in the future, as opposed to reinforcement learning which aims to maximize reward. As stated in the introduction, reinforcement learning can be subsumed within Active Inference since rewards can simply be defined to be highly expected states.

Another difference lies in the representation of the policy. Policy gradient methods optimize the log probability of the policy, while Active Inference directly optimizes the raw probability values. It is still an open question how this changes the dynamics of learning under Active Inference compared to policy gradients, although there is some reason to expect the log-probabilities to be better conditioned. Additionally, Active Inference explicitly computes the integral over the action (at least in discrete action spaces) using the counterfactual predicted results from the action, while the policy gradient method only samples from this integral by using the actions the agent actually took during the episode. This should theoretically reduce variance since it is computing the true expectation rather than the sample advantage, and an analogous scheme also been empirically found to improve performance in actor-critic algorithms \citep*{asadi2017mean,ciosek2018expected}.

The final difference relates to the outer expectation. The policy gradient expression is taken under an expectation over the true environmental dynamics, which are generally unknown. This means that the only way to correctly make updates is to sample the expectation using states that are derived from the current policy. This means that policy gradients are naturally on-policy algorithms which can only validly use the data obtained under the current policy. However, during training the policy changes often which renders past data unusable which decreases sample efficiency. Active Inference, however, requires an expectation taken under the transition model, which is known. This means that Active Inference algorithms can be applied "off-policy", which is a large advantage. Any data can be used to train an Active Inference agent -- even that which was collected under a completely different and perhaps even random policy, provided the transition model is known and accurate. While off-policy variants of policy gradient and actor-critic algorithms have been proposed \citep*{degris2012off,song2015off,haarnoja2018soft} the native off-policy status of Active Inference is a large advantage. 

Interestingly, many of the differences between Active Inference and policy gradients, such as the entropy term and the explicit computation of the expectation over the action, have been theoretically and empirically shown to improve performance of policy gradients. These improvements to policy gradients naturally fall out of the Active Inference framework. The similarities between the two algorithms also highlight a potentially close connection between reinforcement learning and Active Inference, especially the link between maximum entropy reinforcement learning and variational inference. The exact relationships between the paradigms of Active Inference, maximum entropy reinforcement learning, and stochastic optimal control \citep*{friston2011optimal,botvinick2012planning,rawlik2010approximate} still remains unclear, however.
\section{Related Work}

A significant amount of work has gone into exploring tabular Active Inference algorithms and questions such as how the framework encompasses epistemic value and exploration \citep*{friston2015active,pezzulo2016active,moulin2015active,fitzgerald2015active}, models of active vision \citep*{mirza2016scene,friston2018deep,parr2017uncertainty,parr2018active}, biologically plausible neural process theories \citep*{parr2018anatomy,friston2017active}, the connections to the motor system \citep*{adams2013predictions},implementations based on variational message passing \citep*{parr2019neuronal,friston2017graphical,van2019simulating}, and even insight, curiosity, and concept-learning \citep*{friston2017active,schwartenbeck2019computational,smith2019active}. These methods though, while providing insight into the qualitative dynamics of Active Inference and the importance of various parameters, are inherently non-scalable due to their exponential complexity, and they have not been applied to anything beyond simple toy-tasks.

\citet*{ueltzhoffer2018deep}, to our knowledge, is the first paper to propose approximating the observation and transition models with deep neural networks. They use single layer tanh networks with sixteen neurons which outputs the mean and variance of a diagonal conditional gaussian. They used this model to solve the Mountain-Car problem from OpenAI gym. A key difference of this work is how they represented action. They computed continuous actions in a manner that required them to know the partial derivatives of the sensations given the action, which meant propagating through the environmental dynamics, which are unknown. Due to this they had to use a black box evolutionary optimizer to optimize their models, which is substantially more sample-inefficient. In our model we do not use this approach, but instead use a learned amortized inference distribution $Q(a|s)$ and minimize this using a variational approach on the divergence with the approximated true posterior of the value function $p(a|s)$,which is learned through a bootstrapping estimation procedure. Due to this our method is end-to-end differentiable and all networks can be trained through gradient descent on the variational free-energy.

While this paper was in preparation, a paper by \citet*{catal2019bayesian} came out along similar lines. They also parametrized the observation and transition models with deep neural networks, and they used a "habit" policy to approximate the expected free energy, analogously to Q learning in reinforcement learning. However, they only applied their model to the Mountain-Car task and also did not derive the full variational derivation of the KL divergence of the action model and the action posterior, but instead used their habit policy or EFE-approximating network to select actions directly through a softmax choice rule. Instead we maintain a separate policy network which adheres more closely to the full free-energy derivation and also solve significantly more complex tasks than the Mountain-Car.

Active Inference also brings together several contemporary strands of deep reinforcement learning. There has been much work on model-based reinforcement learning which uses models for planning and state estimation, including using separate observation and transition models and unsupervised objectives similar to Active Inference \citep*{deisenroth2011pilco,wayne2018unsupervised,ha2018recurrent}. Deep Active Inference is model-based from the start, and the three separate models effectively fall out of the probabilistic formalism. There has also been work posing the reinforcement learning problem as a variational inference problem. One thread of this work derives maximum entropy reinforcement learning and has been shown to improve benchmark results on many tasks \citep*{rawlik2010approximate,botvinick2012planning,levine2018reinforcement,fellows2018virel,rawlik2013stochastic}. The formalism of Active Inference differs slightly from these in the way it handles rewards and sets up the general MDP formalism. The maximum entropy reinforcement learning inference methods typically sets up the inference problem by assuming binary optimality variables, and then conditions on those variables being true, where the probability of them being true is proportional to the exponentiated reward. Active Inference by contrast does not introduce any auxiliary variables but instead encodes the reward directly in the priors. Beyond this, the detailed connection between Active Inference and the maximum entropy formulation of reinforcement learning remain obscure, despite the fact that they may be equivalent given the close similarities of many of the resulting equations. 

Finally there has also been much work focusing on intrinsic motivations for reinforcement learning agents. For a theoretical review see \citet*{oudeyer2009intrinsic}. There has been work which uses prediction error directly,\citep*{stadie2015incentivizing}, and also information gain \citep*{houthooft2016vime,mohamed2015variational,houthooft2016curiosity} as epistemic rewards in a manner similar to our approach. In Active Inference, however, the form of the epistemic rewards naturally falls out of the framework rather than being postulated on an ad-hoc basis. However, it is still unclear whether the type of epistemic reward proscribed by the expected-free-energy is optimal for exploration, and indeed other forms of epistemic reward may be better. Moreover, it is worth noting, as done in \citep*{biehl2018expanding}, that despite the common use of the expected free energy as the prior, this is in fact arbitrary, and other intrinsic motivations can be substituted.

\section{Model}

All the environments used in this paper were not partially observed, but rather only MDPs with small state-spaces. This means that the mapping from observations to hidden states was unnecessary and thus dispensed with for greater simplicity. The key contribution of this paper is fundamentally the action selection mechanism, and not training a neural network to learn $p(o|s)$. However the transition model was still required and used to compute the epistemic reward.

The policy network, transition network, and value-network were each a two-layered perceptron with 100 hidden units and a relu activation function. All networks were trained through minimizing the free-energy objective using the ADAM optimizer. A learning rate of 0.001 was used throughout. The value network was trained using the bootstrapped estimator. All hyperparameters was kept the same for all tasks. No complex hyperparameter tuning was necessary for reasonable performance on our benchmarks. No preprocessing was done on the states, rewards, or actions. 

The stability of bootstrapped value estimation is a large topic in reinforcement learning. Convergence is not guaranteed for nonlinear function approximators, and stability has been shown to be an issue empirically \citep*{fujimoto2018addressing}. Numerous methods have been discovered in the literature to aid the stability and learning. In this paper we implemented only two of the most basic techniques which are now used universally in deep-Q learning: a memory buffer \citep*{mnih2013playing} and a target network \citep*{van2016deep}. A memory buffer stores a history of previous states visited, and each gradient descent step is taken on a batch of state-action-reward-next-state tuples taken from the buffer. This prevents overfitting to the immediate history, which contains many states that are highly correlated with one another and reduces gradient variance. Secondly, we used a target network, which "freezes" the weights of the value network used in the $\hat{G(s,a)}$ estimator for a number of epochs -- in our experiments we updated after fifty epochs. This enables the value-network to make gradient steps without constantly chasing a moving target, and so aids stability.

Since the bootstrapping estimator for the value function estimator had a significant effect on the behaviour of the model, we believe that large performance gains could be had by implementing many of these techniques and fine-tuning hyperparameters. However, this sort of optimization was not the goal of this paper, which is intended to be more of an introduction and a demonstration of the potential of deep-active-inference rather than a performance contest. Examples of such implementation details can be found in \citep*{fujimoto2018addressing}.

For comparison, we implemented two reinforcement baseline methods: Q-learning and Actor-critic. Q-learning simply learns the state-action value function through a bootstrapping procedure similar to the one we used to approximate the EFE. It does not maintain a separate representation of the policy, but simply chooses actions directly based on the maximum Q-value that it computes. For a fair comparison, the Q-learning agent we implemented used a Boltzmann exploration rule in its action selection, similar to to the softmax over policies implemented in our deep Active Inference algorithm. This also gave the Q-learning agent sufficient stochasticity to explore enough to converge to a good policy for all of the environments.

The Q-learning agent learned a value-network, which was a multi-layer perceptron with a single hidden layer of 100 neurons and a relu activation function. This was identical in the numbers of neurons and activation function to the neural networks used in the deep Active Inference agent. All hyperparameters were kept the same as in the deep Active Inference agent. Like the Active Inference agent, the Q-learning agent used a memory replay buffer and a target network to help stabilize training.

Actor critic algorithms are variants of policy gradient algorithms that use a separate "critic" neural network to estimate the value function instead of directly estimating it through monte-carlo returns. We instantiated the actor critic algorithm with a policy network and a value network identical to the deep-active-inference agent. All hyperparameters were the same as for the deep Active Inference agent. The value-network was trained using Q-learning and also possessed a memory buffer and a target network.

\section{Results}

The performance of the Active Inference and baseline reinforcement learning agents was measured on three tasks from the OpenAI Gym. The tasks were Cartpole Environment, the Acrobot Environment,and the Lunar-Lander environment. While not extremely high dimensional tasks, they are significantly more challenging than any before attempted in the Active Inference literature. Example frames from the three games are shown below.

\begin{figure}[H]
\centering
\begin{subfigure}{.3\linewidth}
    \centering
    \includegraphics[scale=0.3]{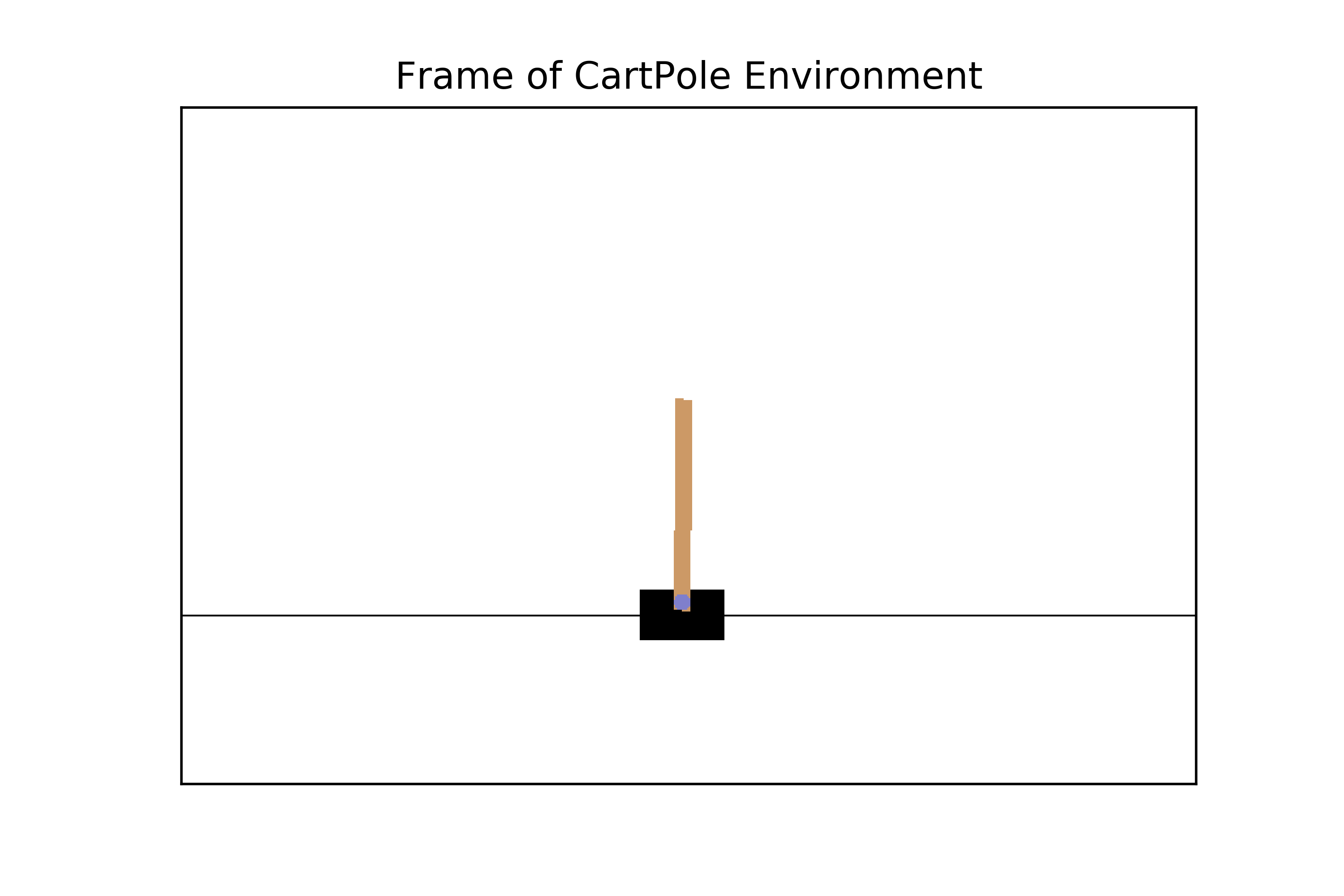}
\end{subfigure}
    \hfill
\begin{subfigure}{.3\linewidth}
    \centering
    \includegraphics[scale=0.35]{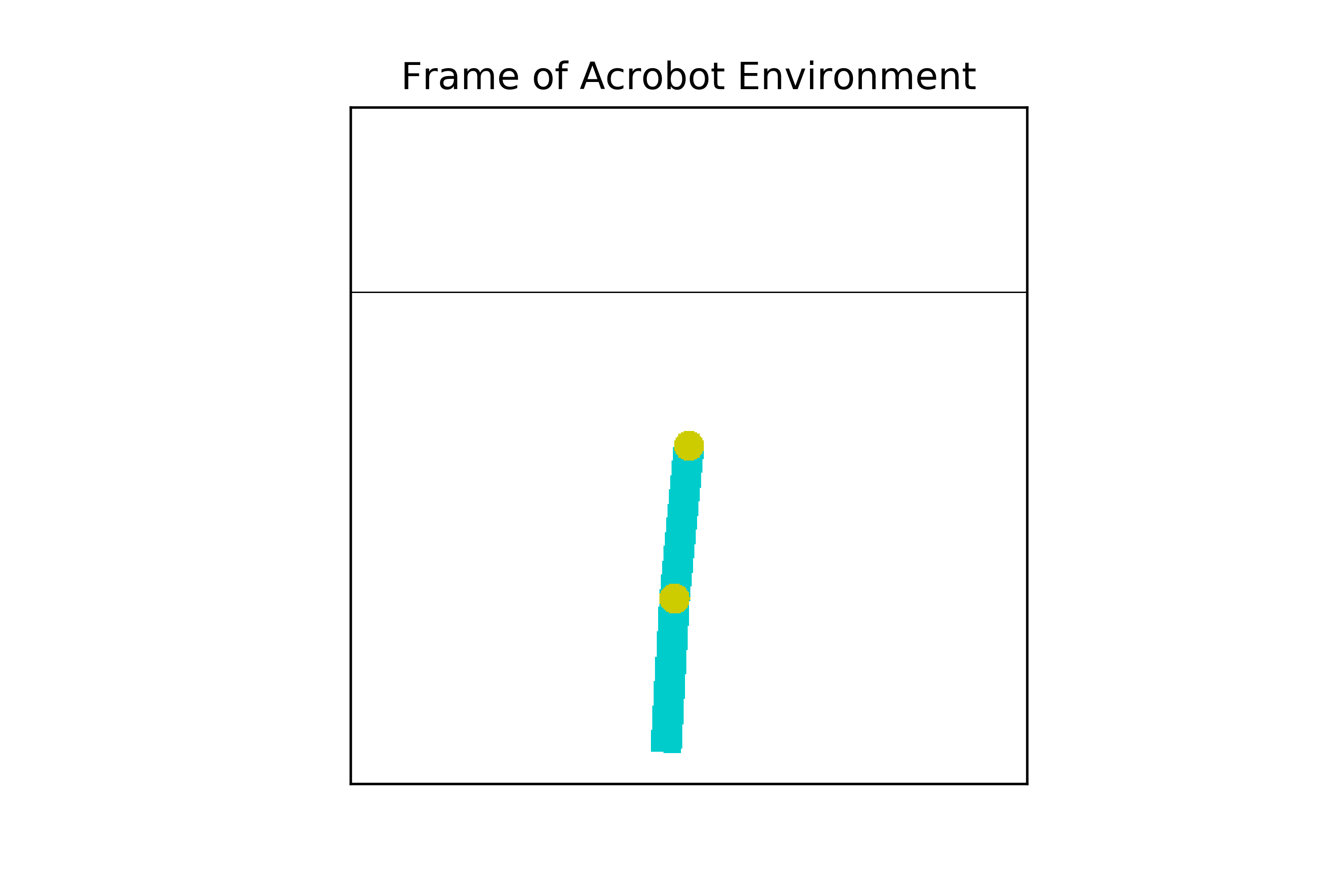}
\end{subfigure}
\hfill
\begin{subfigure}{.3\linewidth}
    \centering
    \includegraphics[scale=0.3]{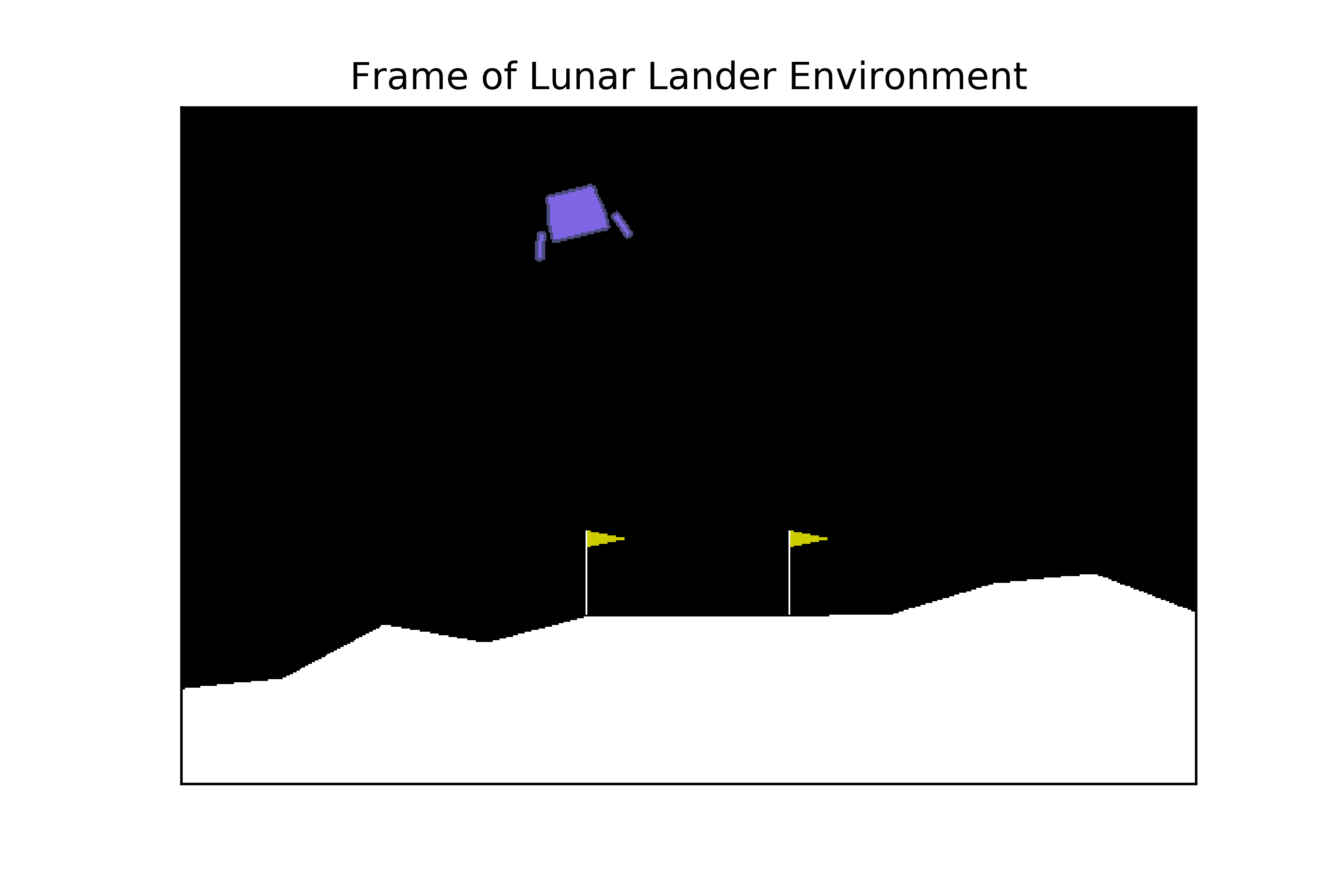}
\end{subfigure}
\RawCaption{\caption{Frames from the three environments. From left to right: CartPole-v1, Acrobot-v1, LunarLander-v2. For more information about these environments beyond what is in this paper, please consult the OpenAIGym documentation.}}
\end{figure}

The goal of the CartPole environment is to keep the pole balanced upright atop the cart. The state space comprises of four values (the cart position and cart velocity, and the angle $\theta$ of the pole and the angle velocity). The reward schedule is +1 for every time-step the episode does not end. The episode ends whenever the angle of the pole is more than 15 degrees from vertical, or the base of the cart moves more than 2.4 units from the center.

In the Acrobot environment, the agent possesses a two jointed pendulum, and the aim is to swing it from a downward position to being balanced vertically at 180 degrees. Reward is 0 if the arm of the Acrobot is above the horizontal and -1 otherwise. This poses a challenging initial exploration problem since getting any reward other than -1 is very unlikely with random actions. The optimal solution would net slightly less than 0 reward (given the time needed to swing up the Acrobot when it would be accruing negative reward). The state-space of the environment is a 6 dimensional vector, which represents the various angles of the joints. The action space is a 3 dimensional vector representing the force on each joint. 

The goal of the Lunar-Lander environment is to land the rocket on a landing pad that is always at coordinates (0,0). The state-space is 8-dimensional and the action space is 4-dimensional with the actions being fire left engine, right engine, upwards engine, and do nothing. The agent receives a reward of +100 for landing on the pad, +10 for each of the rocket-legs are standing, and -0.3 for every time-step the rocket's engines are firing. The maximum possible reward for an episode is +200.

The performance of the Active Inference agent was compared to two baseline reinforcement learning algorithms (Q-learning and actor-critic). Each agent began with randomly initialized neural networks and had to learn how to play from scratch, using only the state and reward data provided by the environment. We ran 20 trials of 15000 episodes each, and the mean reward the agent accumulated on each episode of the CartPole environment is plotted below:

\begin{figure}[H]
    \centering
    \includegraphics[scale=0.6]{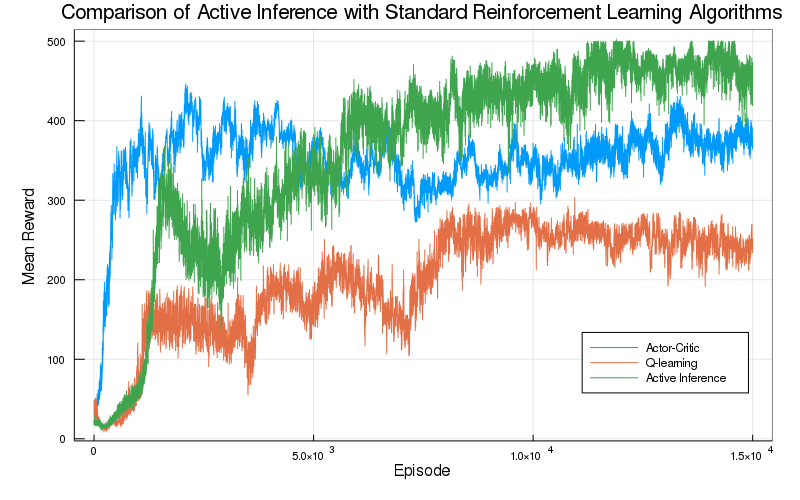}
    \caption{Comparison of the mean reward obtained by the Active Inference agent compared to two reinforcement learning baseline algorithms -- Actor-Critic and Q learning.}
    \label{CartPole Comparison}
\end{figure}

Here we can see that the Active Inference agent actually outperforms both of the reinforcement learning baselines by a significant margin in the end, and the mean reward reaches the maximum score of +500. The actor-critic algorithm does slightly better but does not manage to reach a mean of +500 reward per episode, and the Q learning algorithm performs even worse. This demonstrates that the Active Inference agent can be competitive with, and can even beat conventional reinforcement learning algorithms on some benchmarks.

We now perform an ablation experiment on the Active Inference network to test how the various terms in the algorithm affect performance. We compare the full Active Inference network with two ablated versions. One model lacks the epistemic value component of the value function (see equation 19), and instead estimates the reward only, as in Q learning. The second model lacks the entropy term of the KL loss, and so only optimizes the policy by minimizing $\int Q(a|s)logp(a|s)$ without the entropy term. The results are plotted below:

\begin{figure}[H]
    \centering
    \includegraphics[scale=0.5]{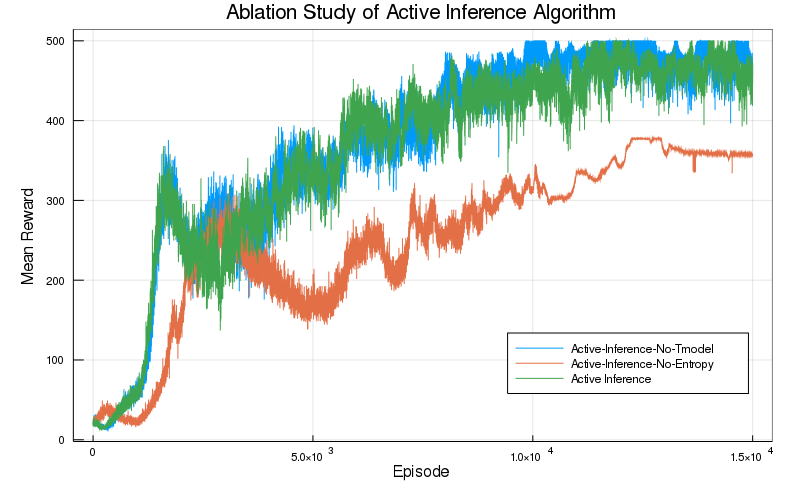}
    \caption{We compare the full Active Inference agent (entropy regularization + transition model) with an Active Inference agent without the transition model, and without both the entropy term and the transition model).}
    \label{Active Inference Ablation}
\end{figure}
An interesting result here is that the main contribution to the success of Active Inference is the entropy term in the loss function. Without the entropy term the Active Inference agent converges to a lower mean reward which is comparable to the performance of the actor-critic and slightly better than the Q-learning algorithm.

\begin{figure}[H]
    \centering
    \includegraphics[scale=0.5]{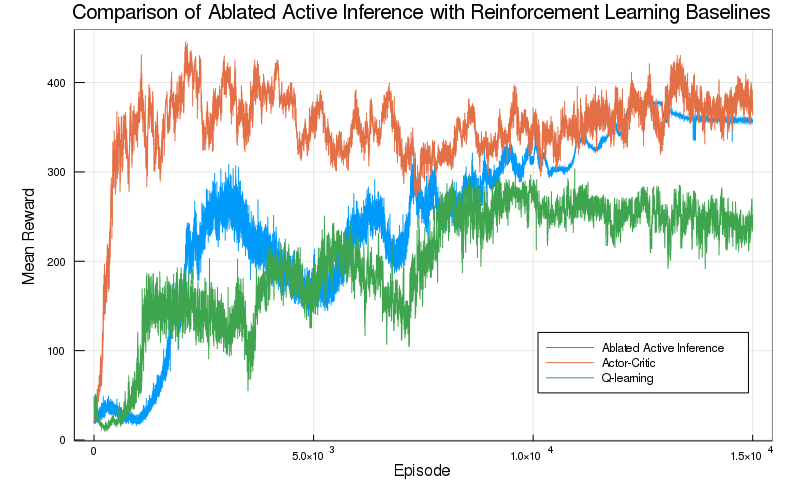}
    \caption{Comparison of the rewards obtained by the fully ablated Active Inference agent with standard reinforcement-learning baselines of Q-learning and Actor-Critic.}
    \label{Ablation Comparison}
\end{figure}

However, the mean rewards are somewhat misleading since the actual distribution of rewards over the runs appears to be bimodal. In most cases all algorithms successfully converged to the maximum of 500 rewards per episode. However, in several cases, the algorithm fails to converge at all, and usually collapses to a very low reward per episode. The mean reward obtained therefore effectively measures the proportion of successful runs rather than the mean of an average run. To see this, the graphs below show a superposition of every single run for the Active Inference agent, the actor-critic, the q-learning agent, and the active-inference-with-entropy agent. We see that the rewards per episode in each run typically bifurcate and either end up being nearly optimal or near zero. This is especially obvious in the ablated Active Inference agent and also in the Q-learning agent, albeit with much more variance. 

\begin{figure}[H]
\centering
\begin{subfigure}{.4\linewidth}
    \centering
    \includegraphics[scale=0.27]{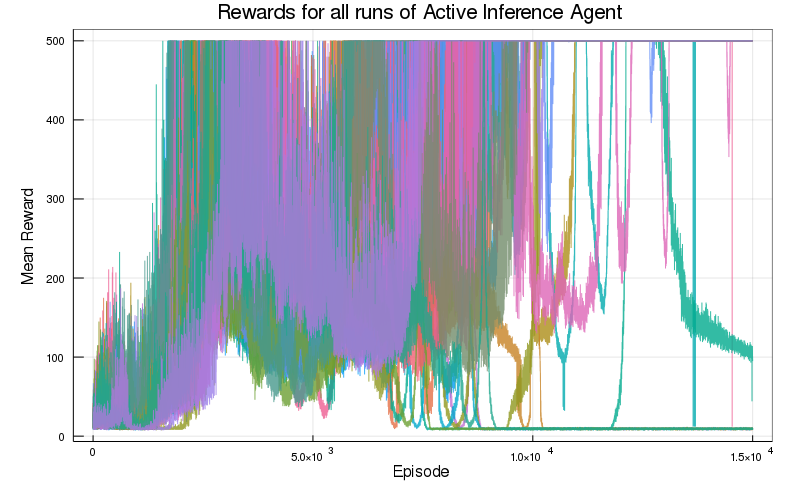}
    \caption{Rewards for all runs of the Ablated Active-Inference Agent.}
\end{subfigure}
    \hfill
\begin{subfigure}{.4\linewidth}
    \centering
    \includegraphics[scale=0.27]{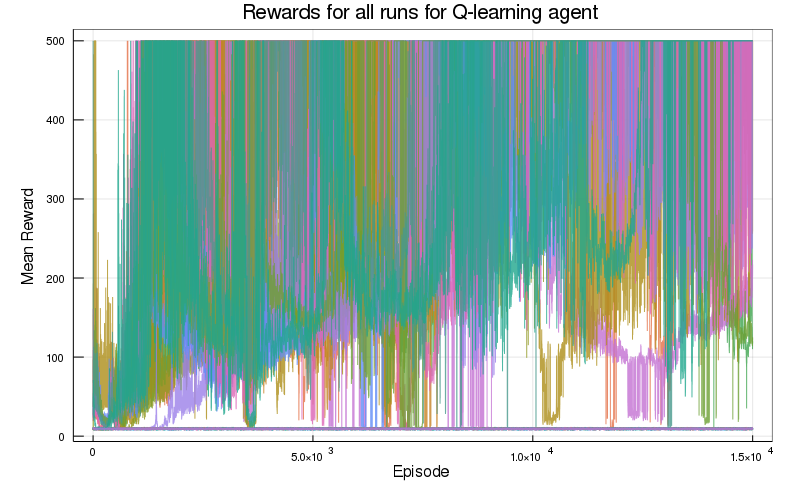}
    \caption{Rewards obtained for all 20 runs of the Q-learning Agent}
\end{subfigure}
\bigskip

\begin{subfigure}{.4\linewidth}
    \centering
    \includegraphics[scale=0.27]{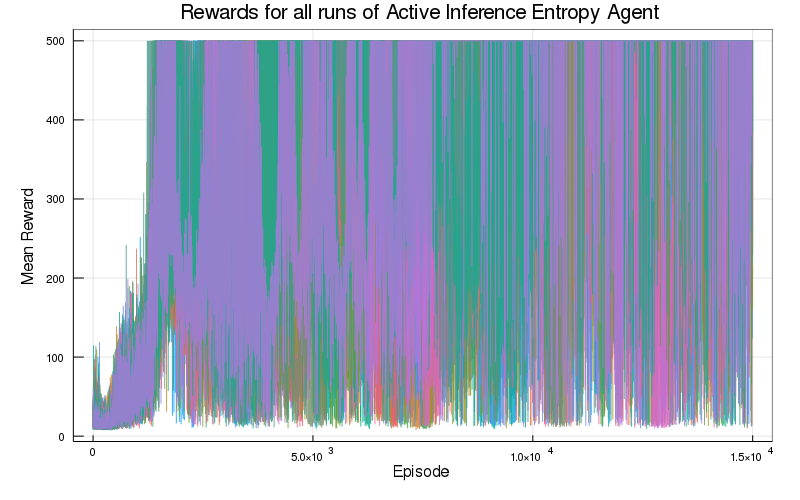}
    \caption{Rewards for all 20 runs of the Active-Inference Agent}
\end{subfigure}
\hfill
\begin{subfigure}{.4\linewidth}
    \centering
    \includegraphics[scale=0.27]{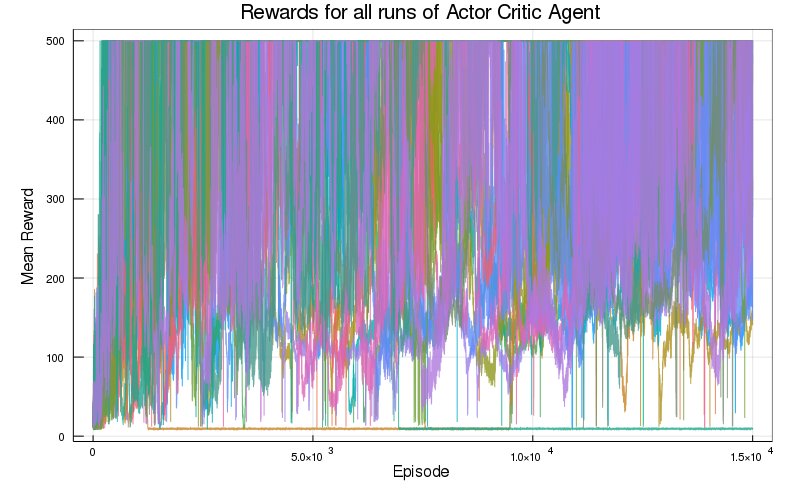}
    \caption{Rewards for all 20 runs of the Actor-Critic Agent}
\end{subfigure}
\RawCaption{\caption{Rewards obtained for each episode for all 20 runs of the different agents. Observe the policy collapse and bifurcations, especially of the Active Inference agent and Q-learning agent for which rewards in an episode will tend towards the optimal +200 or near 0. The entropy term in the Active Inference formulation appears to prevent policy collapse not by causing convergence to a perfect policy, but instead by preventing policy collapses becoming permanent. }}
\end{figure}

We call this bifurcation into either near-optimal or completely failed runs "policy collapse" since, at some point during training, the distribution over actions given by the policy will abruptly collapse to put all weight on a single action, resulting in the agent rapidly tipping over the pole and obtaining a very low score. The entropy-regularized Active Inference algorithm does better because the entropy term encourages the optimizer to spread the probability between all the possible actions as much as possible while also maximizing reward. It is unclear, however, why exactly policy collapse occurs. Moreover, it appears to be a phenomenon the baseline reinforcement learning agents suffer from as well. Interestingly, the entropy term in the Active Inference agent, while it appears to prevent policy collapse, does not simply cause the agents reward to converge cleanly to the maximum. Instead, the reward obtained per episode fluctuates wildly from optimal to near 0, which may be the policy constantly attempting to collapse but being prevented by the entropy term repeatedly.

We found little to no difference when the epistemic value was not computed and the expected-free-energy was simply reduced to the reward. This also held true in the other tasks and runs counter to some of the proposed benefits in the literature for explicit epistemic foraging. We may have observed little effect of the epistemic value for several reasons. Firstly, the tasks used were fairly simple in terms of goals: all they required was simple motor control without any particular need for long-term goal-directed exploration. It is thus possible that random exploration alone, ensured by the entropy-maximizing component of the policy provided sufficient exploration. Secondly, the epistemic value only enters the Active Inference equations as a term in the expected-free-energy, which was estimated through bootstrapping methods. If these estimates were inaccurate or unstable in the first place, then adding an epistemic value term could have little effect. Thirdly, the epistemic value term is defined as the difference between the expected and the observed state posteriors from the transition model. While these were large initially, the magnitude of these prediction errors rapidly declined as the transition model improved, reaching a steady state far smaller than the average extrinsic reward. Thus, the contribution to the expected free-energy from the epistemic rewards would be small, and so would have little impact on behaviour. We demonstrate this by showing the mean time-course of the epistemic reward over the course of the episodes.

\begin{figure}[H]
    \centering
    \includegraphics[scale=0.7]{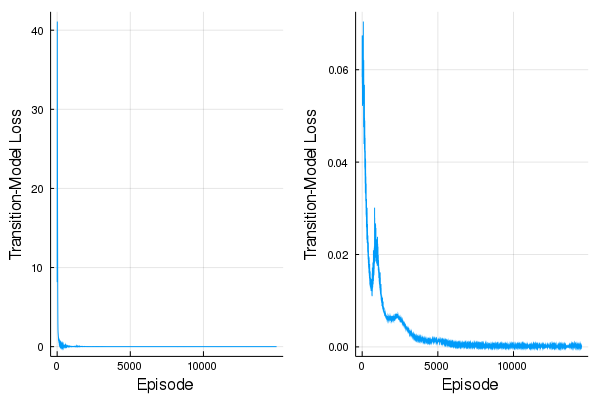}
    \caption{Mean Transition model loss over 15000 episodes. The right graph starts from 500 episodes into a run to show the convergence better since the initial losses are extremely high.}
    \label{Transition Model Graph}
\end{figure}

The transition-model loss declines extremely rapidly to near 0 ,and the agent thus moves from an exploratory to an exploitatory mode. This happens long before the policy or value-networks have converged, meaning that the epistemic value ends up driving very little exploration and having very little effect overall. As a comparison, the reward magnitude of the CartPole task was +1. 

We also compared the Active Inference agent to the two baseline reinforcement learning agents on two more complex tasks than the CartPole -- the Acrobot and the Lunar-Lander environments from OpenAIGym. The graphs of the performance of the agents are below:

\begin{figure}[H]
\centering
    \includegraphics[scale=0.1]{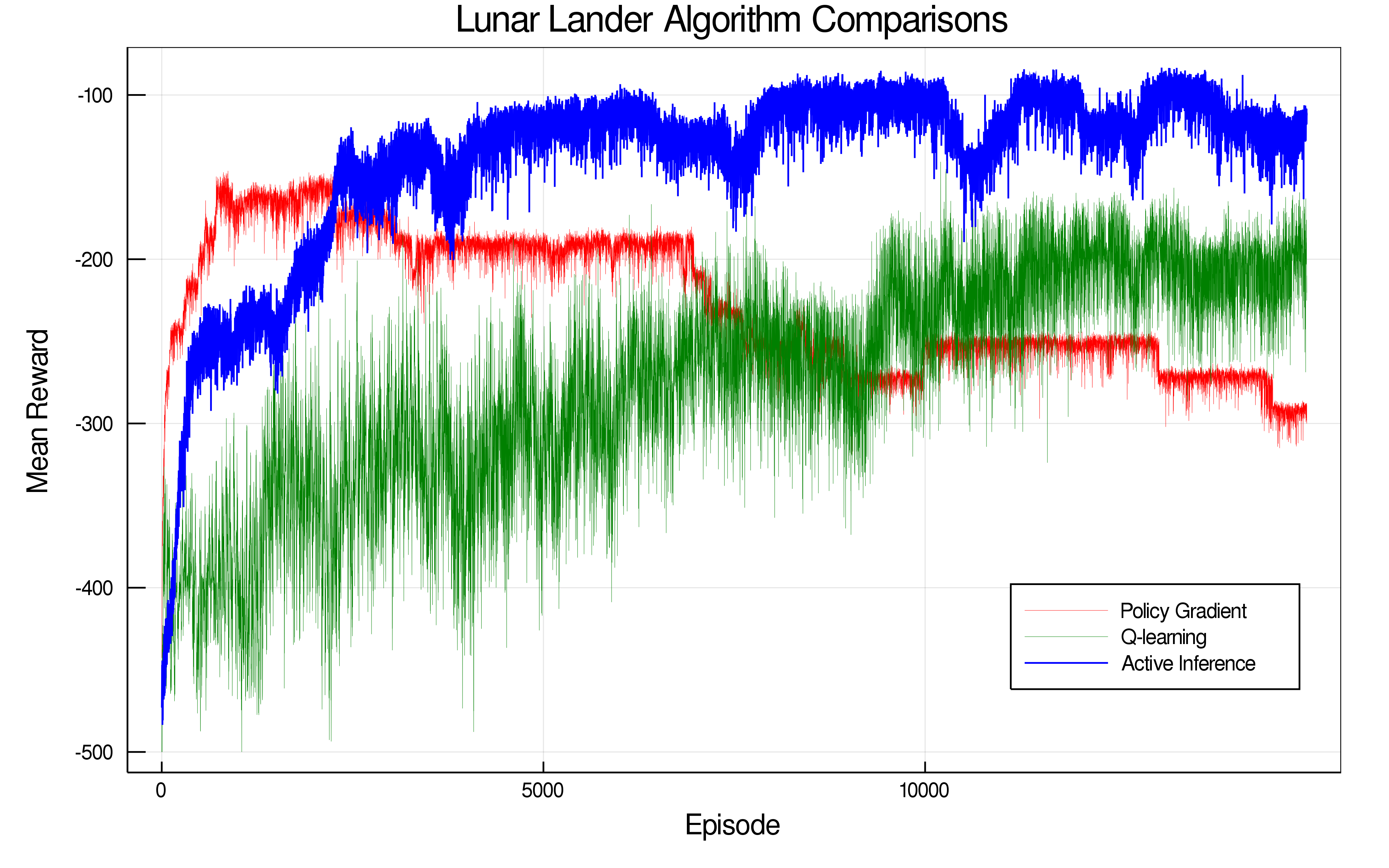}
    \caption{Comparison of Active Inference with standard reinforcement learning algorithms on the Acrobot environment.}
\end{figure}
\bigskip
\begin{figure}[H]
    \centering
    \includegraphics[scale=0.1]{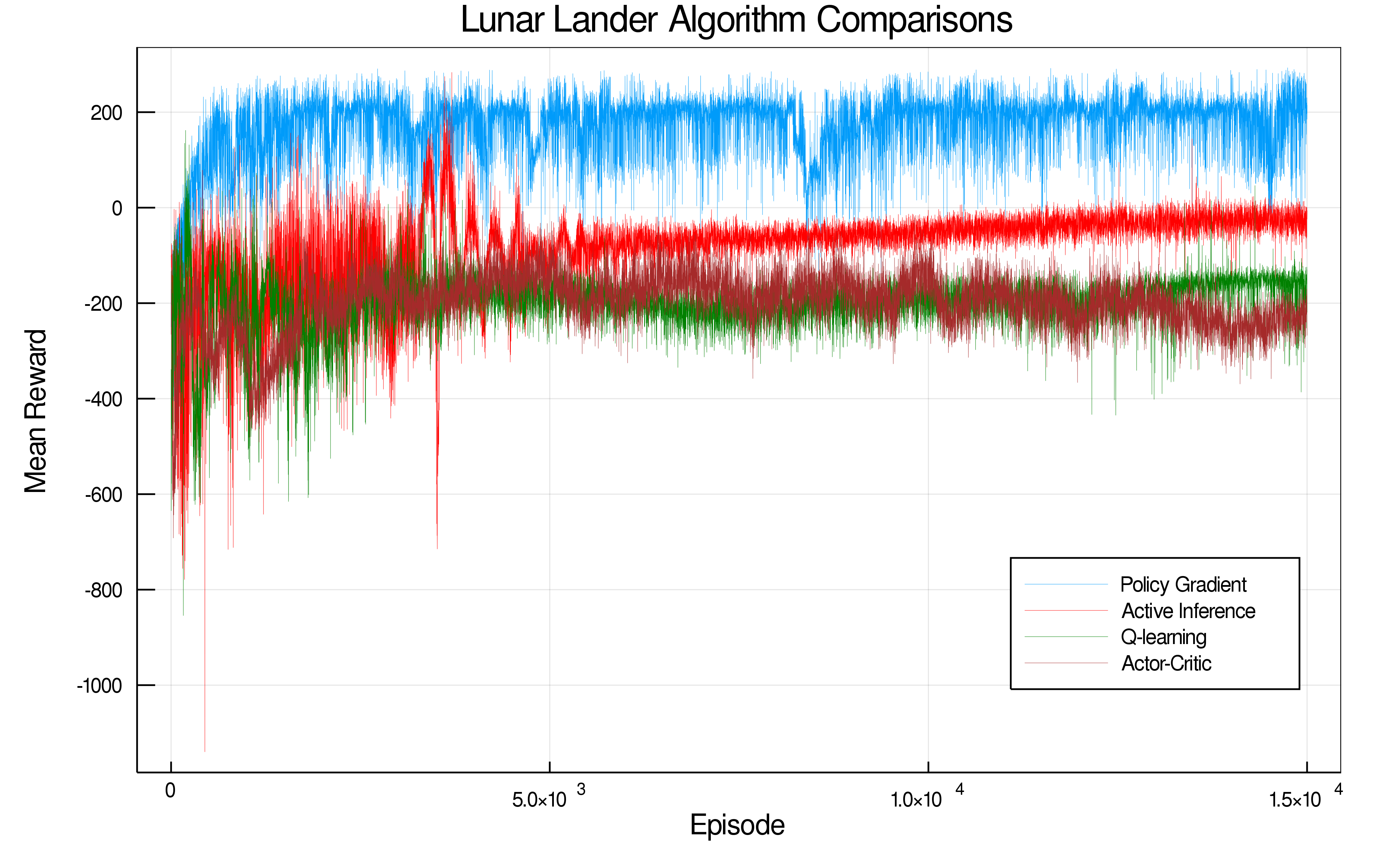}
    \caption{Comparison of Active Inference with reinforcement learning algorithms on the Lunar-Lander environment.}
\end{figure}

As can be seen, the Active Inference agent is competitive with baseline reinforcement leaning agents on both of these tasks. In terms of computational cost, Active Inference without the transition model has roughly the same cost as the deep-actor-critic algorithm from reinforcement learning. Adding the transition model has a greater computational cost and, for the current suite of tasks, appears to add little benefit -- the random exploration ensured by the stochastic action selection and the entropy regularization appears to be sufficient exploration for solving these tasks without more complex epistemic rewards. We believe, however, that in more complex tasks with a compositional structure and long temporal gaps between rewards and actions, then this sort of goal-directed exploration will become increasingly necessary. The Active Inference agent outperforms the two standard reinforcement learning approaches on the acrobot task. This is likely due to the entropy regularization term of the Active Inference agent driving more exploration, which is a key difficulty of this task since no rewards are obtained until the agent happens to swing up the arm to above the horizontal. It is unclear why the performance of policy gradients declines over time in this task, but it could be due to policy collapse. Active Inference underperforms policy gradients on the lunar lander tasks, but it is comparable with Q-learning and actor-critic. We believe this is due to inaccuracies and bias in using neural networks to estimate the value function, as done in actor-critic, Q-learning and our Active Inference algorithm as opposed to simply using unbiased monte-carlo returns, as is done by policy gradient.

\section{Discussion}

In this paper, we have derived a novel deep Active Inference algorithm which uses deep neural networks to approximate the key densities of the variational free energy. We have contrasted this approach with tabular Active Inference and shown that deep Active Inference is significantly more scalable to larger tasks and state-spaces. Moreover, we have shown that our algorithm is competitive, and in some cases superior, to standard baseline reinforcement learning agents on a suite of reinforcement learning benchmark tasks from OpenAIGym. While Active Inference performed worse than direct policy gradients on the Lunar-Lander task, we believe this is due to the inaccuracy of the expected-free-energy-value-function estimation network, since the policy gradient method used direct and unbiased monte-carlo samples of the reward rather than a bootstrapping estimator. Since the performance of Active Inference, at least in the current incarnation, is sensitive to the successful training of the EFE-network, we believe that improvements here could substantially aid performance. Moreover, it is also possible to forego or curtail the use of the bootstrapping estimator and use the generative model to directly estimate future states and the expected-free-energy thereof, at the expense of greater computational cost.

An additional advantage of Active Inference is that due to having the transition model, it is possible to predict future trajectories and rewards N steps into the future instead of just the next time-step. These trajectories can then be sampled from and used to reduce the variance of the bootstrapping estimator, which should work as long as the transition model is accurate. The number N could perhaps even be adaptively updated given the current accuracy of the transition model and the variance of the gradient updates. This is a way of controlling the bias-variance trade-off in the estimator, since the future samples should reduce bias while increasing the variance of the estimate, and also the computational cost for each update. 

Another important parameter in Active Inference and predictive processing is the precision \citep*{feldman2010attention,kanai2015cerebral}, which in Active Inference corresponds to the inverse temperature parameter in the softmax and so controls the stochasticity of action selection. In all simulations reported above we used a fixed precision of 1. However, in tabular Active Inference the precision is often explicitly optimized against the variational free energy, and the same can be done in our deep Active Inference algorithm. In fact, the derivatives of the precision parameter can be computed automatically using automatic differentiation. Determining the impact of precision optimization on the performance of these algorithms is another worthwhile avenue for future work. 

While we did not find that using the epistemic reward helped improve performance on our benchmarks, this could be due to the simplicity of the tasks we were trying to solve, for which random exploration is sufficient. It would be interesting to see if the epistemic value terms of Active Inference become much more important on more complex tasks with a hierarchical and compositional structure, and with long temporal dependencies which are exactly the sort of tasks that current random-exploration reinforcement agents struggle to solve. Moreover, as suggested by \citet*{biehl2018expanding}, Active Inference can also be extended to use other intrinsic motivations, and their effects on the behaviour of Active Inference agents is still unknown.

The entropy regularization term in Active Inference proved to be extremely important, and was often the factor causing the superior performance of Active Inference to the reinforcement learning baselines. This entropy term is interesting, since it parallels similar developments in reinforcement learning, which have also found that adding an entropy term to the standard sum of discounted returns objective improves performance, policy stability and generalizability \citep*{haarnoja2017reinforcement,haarnoja2018acquiring}. This is of even more interest given that these algorithms can be derived from a similar variational framework which also casts control as inference \citep*{levine2018reinforcement}. How these variational frameworks of action relate to one another is an important avenue for future work, particularly since Active Inference possesses a biologically plausible process theory which casts neuronal signalling as variational message passing. Additionally, many of the differences between Active Inference and the standard policy gradients algorithm -- such as the expectation over the action, and the entropy regularization term  -- have been independently proposed to improve policy gradient and actor critic methods. The fact that these improvements fall naturally out of the Active Inference framework could suggest that there is deeper significance both to them and to Active Inference approaches in general. The other key differences between policy gradients and Active Inference is the optimization of the policy probabilities versus the log policy probabilities, and multiplying by the log of the probabilities of the estimated values, rather than the estimated values directly. It is currently unclear precisely how important these differences are to the performance of the algorithm, and their effect on the numerical stability or conditioning of the respective algorithms, and this is also an important avenue for future research. The comparable performance of Active Inference to actor-critic and policy gradient approaches in our results suggest that the effect of these differences may be minor, however.

Our model uses deep neural networks trained using backpropagation, which is generally not thought  to be biologically plausible -- although there are proposed ways to implement backpropagation or approximations thereof in a biologically plausible manner \citep*{whittington2019theories}. This means that our model is not, nor is it intended to be, a direct model of how Active Inference is implemented in the brain. Instead, this work aims towards implementing Active Inference in artificial systems, and providing a proof-of-concept that Active Inference can solve more complex tasks than those currently tackled in the literature and have the potential, ultimately, to be scaled up to the kind of complex problems the brain regularly solves. In terms of biologically plausibility, the variational-message-passing approach of \citet*{parr2019neuronal} seems like a promising direction to take. Our model, however, shows that Active Inference can be applied in a machine learning context using deep neural networks, and can be scaled up to achieve performance comparable with reinforcement learning benchmarks on more complex tasks than any before attempted in the literature. We believe our work takes a step towards answering the question of whether Active Inference approaches can be actually used to solve the kinds of real-world problems that the brain must ultimately solve.

\section{Conclusion}

In sum, we have derived a novel deep Active Inference algorithm directly from the variational free energy. The full model consists of four separate neural networks approximating the terms of the variational free energy functional. We demonstrate that our approach can handle significantly more complex tasks than any previous Active Inference algorithm, and is comparable to common reinforcement learning baselines on a suite of tasks from OpenAIGym. We also highlight interesting connections between our method and policy gradient algorithms and maximum-entropy-reinforcement-learning. Finally, albeit not in a biologically plausible manner, we have shown that Active Inference algorithms can be scaled up to meet large-scale tasks, and that they provide a useful foil to the standard paradigm of reinforcement learning.

\section{Acknowledgements}
I would like to thank Mycah Banks for her invaluable comments and work proofing this manuscript.

\bibliography{AI_PG.bib}

\end{document}